# Reinterpreting 'the Company a Word Keeps': Towards Explainable and Ontologically Grounded Language Models

## Walid S. Saba[*]

*Institute for Experiential AI, Northeastern University, Portland, ME 04101*

## Abstract

We argue that the relative success of large language models (LLMs) is not a reflection on the symbolic vs. subsymbolic debate but a reflection on employing a successful bottom-up strategy of a reverse engineering of language at scale. However, and due to their subsymbolic nature whatever knowledge these systems acquire about language will always be buried in millions of weights none of which is meaningful on its own, rendering such systems utterly unexplainable. Furthermore, and due to their stochastic nature, LLMs will often fail in making the correct inferences in various linguistic contexts that require reasoning in intensional, temporal, or modal contexts. To remedy these shortcomings we suggest employing the same successful bottom-up strategy employed in LLMs but in a symbolic setting, resulting in explainable, language-agnostic, and ontologically grounded language models.

*Keywords:* symbolic embeddings, ontology, compositional semantics, explainability

## 1. Introduction

To arrive at a scientific explanation there are generally two approaches we can adopt, a top-down approach or a bottom-up approach (Salmon, 1989). However, for a top-down approach to work, there must be a set of established general principles that one can start with, which is clearly not the case when it comes to language and how our minds externalize our thoughts in language. In retrospect, therefore, it is not surprising that decades of top-down work in natural language processing (NLP) failed to produce satisfactory results since most of this work was inspired by theories that made questionable assumptions where, for example, an innate universal grammar was assumed (Chomsky, 1957), or that we metaphorically build our linguistic competence based on a set of idealized cognitive models (Lakoff, 1987), or that natural language could be formally described using the tools of formal logic (Montague, 1973). In a similar vein, it is perhaps for the same reason that decades of top-down work in ontology and knowledge representation (Lenat and Guha, 1990; Sowa,

---

[*]w.saba@northeastern.edu

*Draft, last updated May 2, 2024*

1995) also faltered since most of this work amounted to pushing, in a top-down manner, metaphysical theories of how the world is supposedly structured and represented in our minds, and again without any agreed upon general principles to start with. On the other hand, unprecedented progress has been made in only a few years of NLP work that employed a data-driven bottom-up strategy, as exemplified by recent advances in large language models (LLMs) that are essentially a massive experiment of a bottom-up reverse engineering of language at scale (e.g., ChatGPT and GPT-4)[1].

Despite their relative success, however, LLMs do not tell us anything about how language works since these models are not really models of language but are statistical models of regularities found in language[2]. In fact, and due to their subsymbolic nature, whatever 'knowledge' these models acquire about language will always be buried in millions of weights (microfeatures) none of which is meaningful on its own, rendering these models utterly unexplainable (Guizzardia and Guarino, 2024). The fact that the distributed and subsymbolic architectures of neural networks (NNs) are not explainable stems from the following: (i) explainability is 'inference in reverse' but computations in NNs are not invertible; and (ii) distributed and subsymbolic architectures are inherently non-symbolic (van Gelder, 1990) and thus unlike the situation in symbolic systems one cannot in NNs maintain a symbolic structure that preserves a semantic map of the computation, and as summarized by (Browne and Swift, 2020), *there can be no explanation without semantics* (emphasis in original). We summarize this argument of the unexplainability of neural networks (NNs) in figure 1.

Besides unexplainability, LLMs are also oblivious to truth (Borji, 2023), since for LLMs all text (factual or non-factual), is treated equally. Finally, and while LLMs have been shown to do poorly in a number of tasks that require high-level reasoning such as planning (Valmeekam et. al., 2023), analogies (Lewis and Mitchell, 2024) and formal reasoning (Arkoudas, 2023) what concerns here is the failure of LLMs in making the right inferences in various linguistic contexts. As an illustration of the kinds of failures in deep language understanding we consider here three linguistic contexts involving copredication, intension and prepositional attitudes:

**Example 1.** *Show the entities and the relations that are implicit in the following text*: "I threw away the newspaper I was reading because they fired my favorite columnist".

**Example 2.** *Since Madrid is the capital of Spain, can I replace one for the other in the following:* "Maria thinks Madrid was not always the capital of Spain"?

**Example 3.** *Suppose Devon knows that if someone is a client, then s/he is a student, and suppose that Olga is a client.* Then what does Devon know?

The first example involves a phenomenon called *copredication* (see Asher and Pustejovsky, 2005) which occurs when the same entity is used in the same context to refer to more than one semantic

---

[1] GPT stands for 'Generative Pre-trained Transformer', an architecture that OpenAI built on top of the transformer architecture (Vaswani, A. et. al., 2017).

[2] In looking inside the neural network (NN) of an LLM one does not find concepts, meanings, linguistic structures, etc., but weights associated with connections between neurons, which is exactly what one will find in an object recognition or any other NN.

(ontological) type. All LLMs tested[3] failed in recognizing that 'newspaper' in the text is used to simultaneously refer to three entities: (i) the physical object I *threw away*; (ii) the content of the newspaper *I was reading*; and (iii) the 'editorial board' of the newspaper that *did the firing of the columnist*. Note that the failure of the LLMs was more acute when the LLMs were asked to draw a graph showing all entities and relations implied by the text since to show all the relations in the text all the different types of entities must be extracted. Here all LLMs tested showed the same newspaper (*physical*) object doing the firing of the columnist.

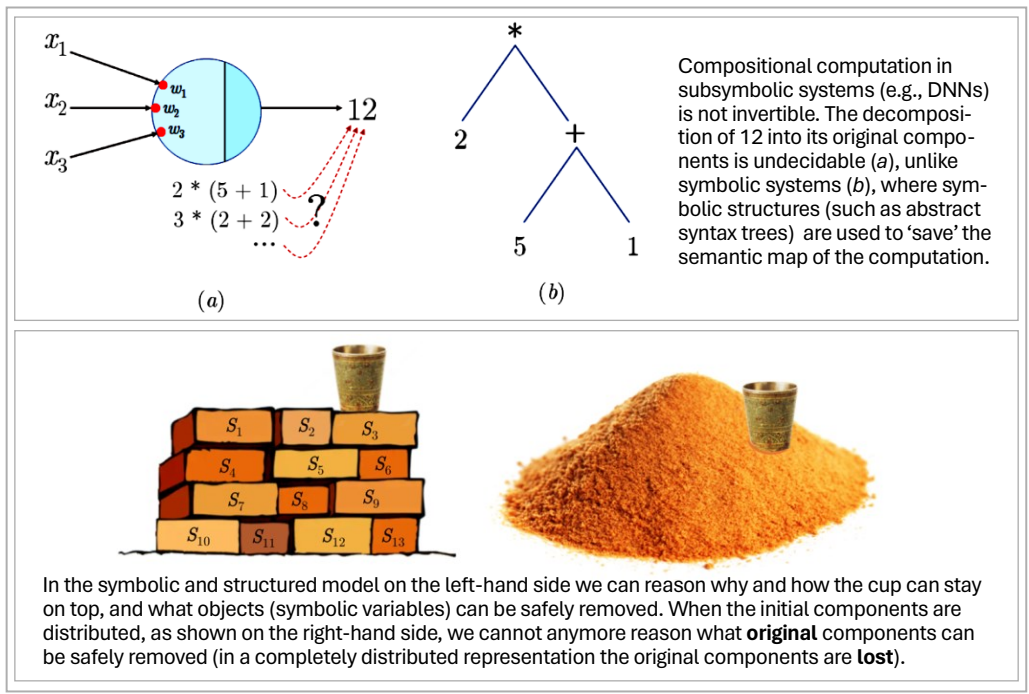

In the symbolic and structured model on the left-hand side we can reason why and how the cup can stay on top, and what objects (symbolic variables) can be safely removed. When the initial components are distributed, as shown on the right-hand side, we cannot anymore reason what **original** components can be safely removed (in a completely distributed representation the original components are **lost**).

**Figure 1.** Explainability is "inference in reverse". Above is a visual explanation of why distributed subsymbolic systems are not explainable since computations in this architecture are not invertible.

In example 2 all LLMs we tested approved replacing 'the capital of Spain' by 'Madrid' resulting in 'Maria thinks that Madrid was not always Madrid'. It is worth noting that the LLMs tested were consistently oblivious to intension. For example, in 'Perhaps Socrates was not the tutor of Alexander the Great', 'Socrates' and 'the tutor of Alexander the Great' were also deemed replaceable (since they are extensionally equal) resulting in 'Perhaps Socrates was not Socrates'. These results were expected since neural networks (deep or otherwise), that are the computing architecture behind all LLMs, are purely extensional models and are based on the 'empiricist theory of abstraction' where their similarity semantics has no notion of 'object identity' (Lopes, 2023). Finally, example 3 illustrates failures of LLMs in making the correct inferences in modal (belief) contexts: the response of the LLMs tested was that 'Devon knows that *Olga* is a student' which is clearly the wrong inference since inferring $\mathbf{K}(Devon, \text{student}(Olga))$ from $\mathbf{K}(Devon, \text{client}(Olga) \supset \text{student}(Olga))$ requires $\mathbf{K}(Devon, \text{client}(Olga))$, i.e., it requires Devon knowing that *Olga* is a client.[4]

---

[3] While our experiments were conducted on several LLMs, GPT-4 (chat.openai.com) and Llama3 (meta.ai) performed best.

So where do we stand now? On one hand, LLMs have clearly proven that one can get a handle on syntax and quite a bit of semantics in a bottom-up reverse engineering of language at scale; yet on the other hand what we have are unexplainable models that do not shed any light on how language actually works. Moreover, it would seem that due to their purely extensional and statistical nature, LLMs will always fail in making the correct inferences in many linguistic contexts. Since we believe the relative success of LLMs is not a reflection on the symbolic vs. subsymbolic debate but is a reflection on a successful bottom-up reverse engineering strategy, we think that combining the advantages of symbolic and ontologically grounded representations with a bottom-up reverse engineering strategy is a worthwhile effort. In fact, the idea that word meaning can be extracted from how words are actually used in language is not exclusive to linguistic work in the empirical tradition, but in fact it can be traced back to Frege.

In the rest of the paper we will (i) first argue that current word embeddings that are the genesis of modern-day large language models can be constructed in a symbolic setting instead of being the result of statistical cooccurrences; (ii) we will show that symbolic vectors perform better than current embeddings on a well-known word similarity benchmark; (iii) we will discuss how our symbolic vectors can be used to 'discover' the ontological structure that seems to be implicit in our ordinary language.

## 2. Concerning the 'Company a Word Keeps'

The genesis of modern LLMs is the *distributional semantics hypothesis* which states that the more semantically similar words are, the more they tend to occur in similar contexts – or, similarity in meaning is similarity in linguistic distribution (Harris, 1954). This is usually summarized by a saying that is attributed to the British linguist John R. Firth that "you shall know a word by the company it keeps". When processing a large corpus, this idea can be used by analyzing co-occurrences and contexts of use to approximate word meanings by word embeddings (vectors), that are essentially points in multidimensional space. Thus, at the root of LLMs is a bottom-up reverse engineering of language strategy where, unlike top-down approaches, "reverse engineers the process and induces semantic representations from contexts of use" (Boleda, 2020). But nothing precludes this idea from being carried out in a symbolic setting. In other words, the 'company a word keeps' can be measured in several ways, other than the correlational and statistical measures that underlie modern word embeddings.

---

[4] For the sake of saving space we refer the reader to other tests we conducted at https://shorturl.at/ejmH8

## 2.1 Symbolic Dimensions of Meaning

In discussing possible models of the world that can be employed in computational linguistics Hobbs (1985) once suggested that there are two alternatives: on one extreme we could attempt building a "correct" theory that would entail a full description of the world, something that would involve quantum physics and all the sciences; on the other hand, we could have a promiscuous *model of the world that is isomorphic to the way we talk it about in natural language* (emphasis is ours). Since the first option is a project that is most likely impossible to complete, what Hobbs is clearly suggesting here is a reverse engineering to discover how we actually use language to talk about the world we live in. This is also not much different from Frege's Context Principal that suggests "never ask for the meaning of words in isolation" (Dummett, 1981) but that a word gets its meanings from analyzing all the contexts in which the word can appear (Milne, 1986). Again, what this suggests is that the meaning of words is embedded (to use a modern terminology) in all the ways we use these words in how we talk about the world. While Hobbs' and Frege's observations might be a bit vague, the proposal put forth by Fred Sommers (1963) was very specific. Again, Sommers suggests that "to know the meaning of a word is to know how to formulate some sentences containing the word" and this would lead, like in Frege's case, to the conclusion that a complete knowledge of some word $w$ would be all the ways $w$ can be used. For Sommers, the process of understanding the meaning of some word $w$ starts by analyzing all the properties $P$ that can **sensibly** be said of $w$. Thus, for example, [*delicious Thursday*] is not sensible while [*delicious apple*] is, regardless of the truth or falsity of the predication. Moreover, and since [*delicious cake*] is also sensible, then there must be a common type (perhaps food?) that subsumes both *apple* and *cake*. This idea is similar to the idea of type checking in strongly typed polymorphic programming languages. For example, the types in an expression such as '$x + 3$' will only unify (or the expression will only 'make sense') if/when $x$ is an object of type number (as opposed to a tuple, for example). As it was suggested in (Saba, 2007), this type of analysis can thus be used to 'discover' the ontology that seems to be implicit in the language, as will be discussed below.

## 2.2 Symbolic Reverse Engineering of Language

The procedure we have in mind assumes a Platonic universe where all concepts, physical or abstract, including states, activities, properties (tropes) (Moltmann, 2013), processes, events, etc. are considered entities that can be defined by a number of language-agnostic primitives (Smith, 2005) that we call the 'dimensions of meaning'. We consider here the following dimensions: AGENTOF, OBJECTOF, HASPROP, INSTATE, PARTOF, INSTATE, INPROCESS, and OFTYPE. For every word $w$ in the language, and for every dimension **D** ∈ {AGENTOF, OBJECTOF, HASPROP, INSTATE, PARTOF, INSTATE, INPROCESS, and OFTYPE }, a reverse-engineering process is conducted by computing a set $w^{\mathbf{D}}$ that is a set of pairs $(x, t)$ and where $t$ is a weight in [0,1]. Here are example sets computed for 'book' across four dimensions of meaning along with the corresponding masking prompt that queries what some LLM has 'learned' about how we talk about books in ordinary language:

book . HASPROP
*Everyone likes to read a* [**MASK**] *book.*
=> {(popular, 0.9), (educational, 0.8), (famous, 0.8), ... }

book . OBJECTOF
*Everyone I know enjoyed* [**MASK**] *'The Prince'.*
=> {(reading, 0.9), (writing, 0.8), (editing, 0.8), ... }

book . AGENTOF
*Das Kapital has* [**MASK**] *many people over the years.*
=> {(influenced, 0.9), (inspired, 0.8), (changed, 0.8), ... }

book . PARTOF
*Hamlet should be part of every* [**MASK**].
=> {(collection, 0.9), (archive, 0.8), (library, 0.8), ... }

book . INSTATE
*I was told that my book is not in* [**MASK**].
=> {(print, 0.9), (circulation, 0.8), (review, 0.8), ... }

What the above says is the following (i) in ordinary spoken language we speak of a 'book' that is *popular*, *educational*, *famous*, etc.; (ii) we speak of *reading*, *writing*, *editing*, etc. a 'book'; (iii) we speak of a 'book' that may *change*, *influence*, *inspire*, etc.; (iv) we speak of a b 'book' that is part of a *collection*, an *archive*, or a *library*; and (v) a book can be in *review*, in *print*, in *circulation*, etc. Note that members of these sets are themselves nominalized into concepts that are properties, activities, processes, etc. and that in turn have their own computed sets, as shown in figure 1 where a reading (an object of type activity) that is an element of book . OBJECTOF, has its own web of concepts and across the same dimensions of meaning.

The nominalization process can be conducted using the copular 'is' as shown in table 1. For example, 'John is famous' can be restated as 'John *has the property* of fame'; 'Jim is sad' as 'Jim is *in a state* of sadness'; etc. (see [Smith, 2005] for details on the relationship between the copular and abstract entities and [Moltmann, 2013] for more on abstract objects.)

### 2.3 Symbolic Embeddings

The process we described thus far would result in symbolic word embeddings as the one shown in figure 2 below. In figure 2(a) we show the symbolic embedding for 'boy' and 'lad' along the HASPROP dimension. Thus, in ordinary spoken language it is sensible to speak of a 'handsome boy' and a 'funny boy' as well as a 'clever lad' and a 'talented lad'. We note here that in this process generic descriptions are removed using a function that computes the information content of some adjectives, where the information content of an adjective *adj* is inversely proportional to the set of types of *adj* can sensibly be applied to. For example, 'beautiful' will have a low information content score since 'beautiful' can sensibly be said of many concepts, both physical and abstract (e.g., *car*, *movie*, *poem*, *night*, *girl*, ...) while 'tasty' can sensibly be said of 'food' and just a few others. The symbolic embeddings in figure 2(b) are those of 'automobile' and 'car' along the OBJECTOF dimension. Note now that word similarity along these symbolic dimensions can be computed using cosine similarity as well as weighted Jaccard similarity where max and min can be used in fuzzy union and fuzzy intersection.

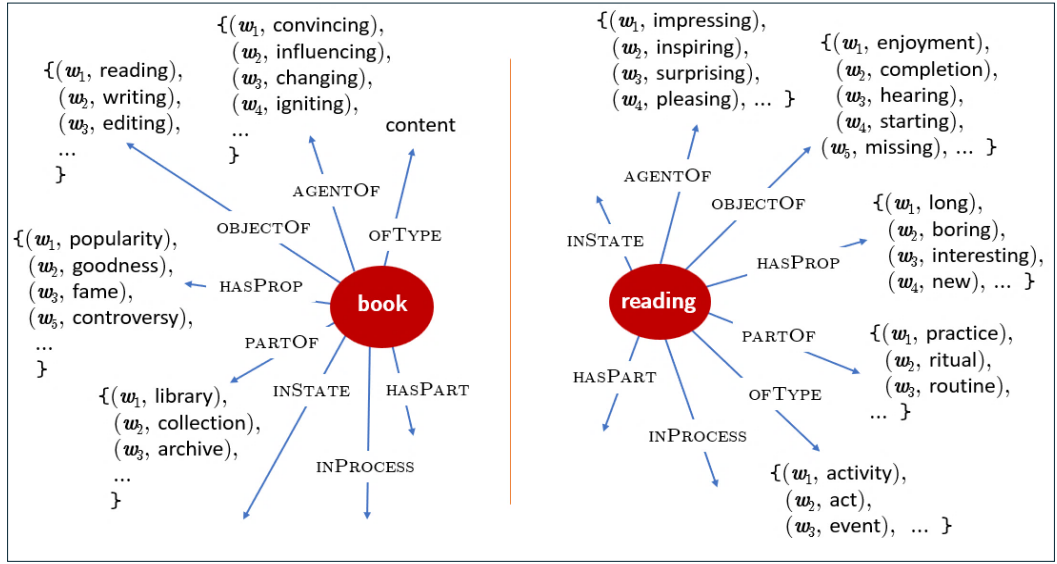

**Figure 1.** The dimensions of meaning of 'book' and 'reading'. Note that the 'reading' activity is one of the elements of the set of concepts that a 'book' can be the OBJECTOF.

Table 1. Nominalization of copular sentences to discover dimensions of meaning

| Proposition | Nominalized Form (entities and relations) |
|---|---|
| John is famous | (*John* : person) —— HASPROP —— (*fame* : property) |
| Jim is sad | (*Jim* : person) —— INSTATE —— (*sadness* : state) |
| Maria is recognized | (*Maria* : person) —— OBJECTOF —— (*recognition* : event) |
| Olga is dancing | (*Olga* : person) —— AGENTOF —— (*dancing* : activity) |
| Sara is maturing | (*John* : person) —— GTPROCESS —— (*maturity* : process) |
| Hamlet is inspiring | (*Hamlet* : book) —— AGENTOF —— (*inspiration* : act) |
| Fame is desirable | (*fame* : property) —— HASPROP —— (*desirability* : property) |
| Death is inevitable | (*death* : state) —— HASPROP —— (*inevitability* : property) |

The final similarity could then be a weighted average function **f** of the similarities across the seven dimensions of meaning:

$$\textbf{sim}(\text{`boy'}, \text{`lad'}) = \textbf{f}\left(\{\textbf{dimensionSim}^D(\text{`boy'}, \text{`lad'}) \mid D \in \textit{dims}\}\right)$$

where

$$\textit{dims} = \{\textsc{hasProp}, \textsc{agentOf}, \textsc{objectOf}, \textsc{inState}, \textsc{ofType}, \textsc{hasProp}, \textsc{inProcess}\}$$

We are currently experimenting with the optimal number of dimensions using a number of word similarity benchmarks, including the WordSim353 dataset (Finkelstein, Lev, et al., 2001)[5].

What should be noted here is that even with the simple conceptual structure discovered thus far one can, given a prompt, generate plausible text, such as the following text fragments:

(1)  *I enjoyed the interesting reading of the new book.*
(2)  *They completed the boring reading of the controversial book.*

The sensible (and meaningful) fragment in (1) can be generated because a *book* can be 'read' and can be described by 'new', and *readings* can be 'interesting' and they can be the object of *enjoyment*; and similarly for (2) where a *reading* of a *controversial* book can be *boring* and can be the object of a *completion*, etc. Note, however, that text generation in this case is not a function of 'predicting' the most likely continuation, but a function of plausible filling in of subjects, objects, agents, descriptions, etc. to any propositional structure.

| a very [MASK] boy | a very [MASK] lad |
|---|---|
| (handsome, 0.93) | (clever, 0.91) |
| (cute, 0.91) | (handsome, 0.90) |
| (naughty, 0.89) | (nice, 0.88) |
| (nice, 0.87) | (clever, 0.86) |
| (clever, 0.85) | (adorable, 0.85) |
| (pretty, 0.81) | (polite, 0.81) |
| (funny, 0.78) | (funny, 0.79) |
| (talented, 0.77) | (pretty, 0.76) |

(a)

| I was [MASK] an automobile | I was [MASK] a car |
|---|---|
| (driving, 0.93) | (driving, 0.93) |
| (riding, 0.93) | (manufacturing, 0.93) |
| (manufacturing, 0.93) | (chasing, 0.93) |
| (repairing, 0.93) | (riding, 0.93) |
| (owning, 0.93) | (restoring, 0.93) |
| (leasing, 0.93) | (owning, 0.93) |
| (designing, 0.93) | (leasing, 0.93) |
| (chasing, 0.93) | (buying, 0.93) |

(b)

**Figure 2**. vectors for 'boy' and 'lad' across the HASPROP dimension (a), and vectors for 'automobile' and 'car' across the OBJECTOF dimension (b).

---

[5] https://www.kaggle.com/datasets/julianschelb/wordsim353-crowd

## 3. The Ontology of the Language of Thought

The reverse engineering process we have described above would result in symbolic embeddings along various dimensions, as the ones shown in figure 2. As a result of this, however, we could then analyze the subset relations between these embeddings to discover the ontological structure that seems to be implicit in our ordinary language. To illustrate, consider the following:

(3)   car . OBJECTOF = {(driving, 0.9), (repairing, 0.8), (buying, 0.8), ... }
(4)   book . OBJECTOF = {(reading, 0.9), (writing, 0.8), (buying, 0.8), (selling, 0.8), ... }
(5)   person . AGENTOF = {(reading, 0.9),(writing, 0.8), (driving, 0.8), (buying, 0.8), ... }
(6)   person . HASPROP = {(popularity, 0.9), (fame, 0.8), ... }
(7)   car . HASPROP = {( popularity, 0.9), (fame, 0.8), (beautiful, 0.8), ... }
(8)   book . HASPROP = {(popularity, 0.9), (fame, 0.8), (beautiful, 0.8), ... }

Note that car can be the object of 'buying' and so can be a book and this means that car and book must, at some level of abstraction, share the same parent (perhaps 'artifact'?) Note also that a car as well as a book and a person can be popular. An analysis along these lines would result in the following, where $P(t_1)$ means P applies to objects of type $t_1$, and $Q(t_1, t_2)$ means that Q is a relation that holds between objects of type $t_1$ and $t_2$:

(9)    READ(person, book)
(10)   WRITE(person, book)
(11)   BUY(person, $T_1$ = car $\cup$ book ... )
(12)   DRIVE(person, car)
(13)   BEAUTIFUL($T_2$ = person $\cup$ car $\cup$ book ... )

What the above says is the following: in ordinary spoken language we speak of people reading and writing books (9 and 10); we speak of people buying cars and books, and thus of buying objects that are of some type that subsumes both cars and books (11); we speak of people driving cars (12); and we speak of beautiful people, cars, and books (and thus beautiful seems to be a property that can sensibly be said of concepts that are at very high level of generality). The subtyping relationships are then determined using these type constraints. For example, consider the following:

(14)   MANUFACTURE(person, computer)
(15)   MANUFACTURE(person, car)
(16)   MANUFACTURE(person, couch)
(17)   ASSEMBLE(person, computer)
(18)   ASSEMBLE(person, car)
(19)   ASSEMBLE(person, couch)

That is, it makes sense to say (or it is sensible to speak of) manufacturing and assembling a computer, a car, and a couch. This means that computer, car, and couch must have a common parent at some point of generalization (perhaps artifact?). However, while the following are also sensible,

(20)   IDLE(computer)
(21)   IDLE(car)
(22)   ON/OFF(computer)
(23)   ON/OFF(car)

that is, since we can speak of an idle computer and an idle car, as well as a computer and a car being an ON/OFF, while the same is not true of a couch, then car and computer must have a more specific type than artifact that they do not share with couch, as shown in figure 3. As suggested by Sommers (1963) this type of analysis that can be fully automated with the help of LLMs can help us discover what he called 'the Tree of Language' – which is essentially the ontology that seems to be underneath our ordinary language. This might also be what Hobbs (1985) was seeking when he suggested building a model of the world that isomorphic to the we talk about it in natural language.

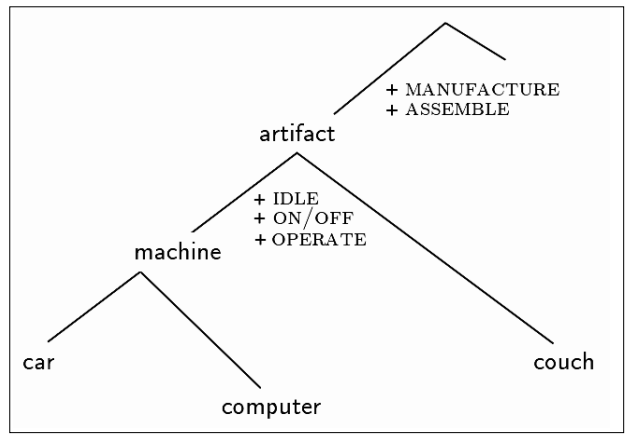

**Figure 3.** We manufacture and assemble cars, computers, and couches and thus they must all share a common parent at some level of abstraction, while car and computer differ from couch in that we speak of operating both and of both being idle or on/off while the same is not true of couches.

How we can use the discovery of the ontological structure that seems to lie underneath ordinary language in reasoning is beyond the scope of this paper.

## 4. Concluding Remarks

Large language models (LLMs) have shown impressive capabilities that pioneers in artificial intelligence and natural language processing would marvel at. However, we believe that LLMs are not the answer to the language understanding problem nor to reasoning in general and in particular commonsense reasoning. Due to their paradigmatic unexplainability LLMs will also not shed any light on how language works and how we externalize our thoughts in language. Since, in our opinion, the relative success of LLMs is not due to their subsymbolic nature but due to applying a successful bottom-up reverse engineering strategy, we suggested here applying the same strategy but in a symbolic setting, something that has been argued for by logicians dating back to Frege. By combining the successful bottom-up strategy and symbolic and ontological methods we arrive at explainable and ontologically grounded language models that can be used in problems requiring commonsense reasoning.

We are still in the early stage of this work, but we currently have the tools to realize the dream of Frege and Sommers and perhaps shed some light on the 'language of thought' (Fodor, 1988) – the internal language that we use to construct and process our thoughts.